\documentclass[a4paper,fleqn]{cas-dc}

\usepackage[numbers]{natbib}

\usepackage{graphics} 
\usepackage{epsfig}
\usepackage{amsmath} 
\usepackage{amssymb}  
\usepackage{balance}
\usepackage{color}
\usepackage{soul}
\soulregister\cite7
\soulregister\ref7

\usepackage{caption}
\usepackage{subcaption}
\usepackage{changepage}

\newcommand{\lf}[1]{{}^{\scriptscriptstyle {#1}}\hspace{-0.1mm}}
\newcommand{\lfR}[1]{{}^{\scriptscriptstyle {#1}}\hspace{-0.8mm}R}

\newcommand{\rf}[1]{_{\scriptscriptstyle {#1}}\hspace{-0.0mm}}
\newcommand{\lfdotR}[1]{{}^{\scriptscriptstyle {#1}}\hspace{-0.8mm}\dot{R}}

\newcommand{\vect}[1]{\boldsymbol{\mathbf{#1}}} 

\DeclareMathOperator*{\Exp}{\mathrm{Exp}}
\DeclareMathOperator*{\Ln}{\mathrm{Ln}}

\def\tsc#1{\csdef{#1}{\textsc{\lowercase{#1}}\xspace}}
\tsc{WGM}
\tsc{QE}

\begin{document}
\let\WriteBookmarks\relax
\def\floatpagepagefraction{1}
\def\textpagefraction{.001}

\newcommand{\bac}{Best Axes Composition Extended: Multiple Gyroscopes and Accelerometers Data Fusion to Reduce Systematic Error}

\shorttitle{\bac{}}    

\shortauthors{Marsel Faizullin and Gonzalo Ferrer}  

\title [mode = title]{\bac{}}  



%

\author[]{Marsel Faizullin}[
    orcid=0000-0002-1053-7771
]



\ead{marsel.faizullin@skoltech.ru}



\affiliation[]{organization={Skolkovo Institute of Science and Technology},
            city={Moscow},
            country={Russia}}

\author[]{Gonzalo Ferrer}[
    orcid=0000-0003-2704-7186
]


\ead{g.ferrer@skoltech.ru}







\begin{abstract}
Multiple rigidly attached Inertial Measurement Unit (IMU) sensors provide a richer flow of data compared to a single IMU. 
State-of-the-art methods follow a probabilistic model of IMU measurements based on the random nature of errors combined under a Bayesian framework.
However, affordable low-grade IMUs, in addition, suffer from systematic errors due to their imperfections not covered by their corresponding probabilistic model.
In this paper, we propose a method, the Best Axes Composition (BAC) of combining Multiple IMU (MIMU) sensors data for accurate 3D-pose estimation that takes into account both random and systematic errors by dynamically choosing the best IMU axes from the set of all available axes.
We evaluate our approach on our MIMU visual-inertial sensor and compare the performance of the method with a purely probabilistic state-of-the-art approach of MIMU data fusion. We show that BAC outperforms the latter and achieves up to 20\% accuracy improvement for both orientation and position estimation in open loop, but needs proper treatment to keep the obtained gain. 
\end{abstract}




\maketitle

\section{Introduction}\label{sec_intro}

Pose estimation is widely used nowadays for aviation, camera stabilization, virtual reality, visual-inertial odometry, etc. To achieve this, Inertial Measurement Unit (IMU) sensors are key to provide information regarding the angular velocity and acceleration of a moving object at a high-frequency. These kinds of sensors have become ubiquitous with modern and cheap manufacturing techniques, and one can find them integrated into cameras, smartphones, and other electronic devices, sometimes even multiple of them.
Our aim is to provide a sensor fusion technique that combines multiple IMUs, where the most common task is the estimation of position and orientation, either as a standalone task for the IMUs, also known as dead-reckoning, or in conjunction with a complementary sensor, such as a camera, in what is known as visual-inertial methods.


The mass adoption of IMUs is due to the success of 
Microelectromechanical systems (MEMS), in particular MEMS IMU. This technology results in affordable sensors at the cost of limited quality. Multiple of these sensors should improve their accuracy to any downstream task if appropriately combined. 
In this paper, we study data fusion of MEMS IMUs, whose data generation distributions do not follow the common models (found in better quality IMU). Effects such as non-linearity~\cite{wang2019optimized}, sensor finite dimensions~\cite{rehder2016extending}, discrete integration, non-Gaussian distributions, non-stationary parameters, temperature dependence or power supply conditions are denoted as {\em systematic} errors or {\em epistemic} errors.

Our proposed method, rather than following a Bayesian combination of data, monitors the systematic error and dynamically identifies and selects the best fitting data from each sensor axis, and discards the rest due to the higher implicit systematic errors. Our contribution, named {\em Best Axes Composition} (BAC) follows the hypothesis that it is {\em better to keep the information gathered from the source with the minimal systematic error rather than use every piece of data from all sensors}, as the classical Bayesian approach dictates.
Rejecting data can be considered as statistical hypothesis testing, where the most accurate and precise distribution of data is taken into account while other distributions are simply categorized as outlier distributions.
Well-known examples of this approach include RANSAC~\cite{fischler1981random} or the  Minimum Filter~\cite{rfc1059} used in some versions of the Network Time Protocol (NTP).

We will show the advantages of BAC empirically compared to  (i) single IMU and (ii) Multiple IMU (MIMU) Averaged Virtual Estimator (AVE). Single IMU data along with visual information provide comprehensive results in visual-inertial odometry problem~\cite{li2013high, forster2016manifold}. MIMU AVE can outperform single IMU methods and shows better accuracy of trajectory estimation~\cite{zhang2019lightweight, bancroft2011data, wang2015}. 

The present document is an extension of the gyroscope Best Axes Composition (BAC-gyroscope) introduced in~\cite{faizullin2021best}. This new material includes the accelerometer Best Axes Composition (BAC-accelerometer) to complete our approach for inertial data and for the entire pose estimation in open loop. Although both variants of BAC are formally similar, the obtained results suggest that the ranking of the systematic error distributions in the BAC-gyroscope remains valid over longer periods of time (3-4s), while BAC-accelerometer improves accuracy for much shorter time windows (0.5s).

The paper is organized as follows: 
in Sec.~\ref{sec_rl}, the current research on MIMU data fusion is discussed; 
Sec.~\ref{sec_background} contains the introduction of IMU and kinematic models utilized in this work;
Sec.~\ref{sec_approach} is aimed at the description of the proposed method; 
Sec.~\ref{sec_impl} introduces a visual-inertial sensor exploited in the experiments;
Sec.~\ref{sec_exp} covers methodology of the experiments and experimental results;
Sec.~\ref{sec_concl} is for the conclusion. 

\section{Related Work}\label{sec_rl}

The hypothesis that measurements from multiple sensors from the same kind can improve the accuracy of pose estimation is considered as a gold standard due to the increase of information in comparison to a single sensor. To examine this hypothesis, the application of rigidly-connected MIMU sensors has been investigated for decades.
In the work of Nilsson \textit{et al.}~\cite{nilsson2016inertial}, the authors count more than three hundred publications on methods of MIMU data fusion. 
Among the reasons for applying MIMU sensor-based methods, one can list the following ones:
higher accuracy, higher reliability, more accurate uncertainty estimation, higher dynamic measurement range, estimation of angular motion from accelerometer data, direct estimation of angular acceleration, etc. 

MIMU data fusion goes alongside the term Virtual IMU (VIMU). This kind of a single pseudo-IMU substitutes several real IMUs by its own measurements and frame to be used for estimation algorithms~\cite{bancroft2011data}. The introduction of VIMU is reasonable due to the well-defined kinematics of rigidly connected frames expressed by kinematic equations~\cite{kevin2017modern}.

Skog~\textit{et al.}~\cite{skog2014open} introduce an open-source MIMU platform and discuss possible applications of it. One of the usages they state is stochastic error reduction by averaging out independent errors of every sensor since the averaged measurements have theoretically lower variance.
The authors also discuss the possibility of redundant MIMU measurements to handle the failure of one of the sensors while the others operate normally. 
The authors reason that a specific way of fusing data from sensors with different dynamic ranges can (i) extend the overall dynamic range and (ii) make measurements more accurate thanks to sensors with lower dynamic range have lower noise density while others have a high dynamic range, though, with the increased value of noise. 
Skog~\textit{et al.}~\cite{skog2016inertial}~exploit geometrical constraints between rigidly connected IMUs on an MIMU platform to obtain constraints between the kinematics of every IMU: they employ an array of accelerometers to estimate angular velocities of the platform. Liu~\textit{et al.}~\cite{liu2014design}~and~Edwan~\textit{et al.}~\cite{edwan2011constrained} apply methods for angular velocity estimation by only four triaxial and 12 separated monoaxial accelerometers data respectively. Zhang~\textit{et al.}~\cite{zhang2019lightweight} describe theoretical aspects of up to 9 IMU sensors data fusion technique and show accuracy improvement against a single IMU approach. 

Rhudy \textit{et al.}~\cite{rhudy2012fusion} describe the use of three EKF-based fusion methods of GPS and redundant IMU data while redundant IMU data are processed by: (a) averaging of IMU data before filtering, (b) predicting state based on separate IMU measurements then updating the averaged state and (c) fully separate filtering with averaging of the obtained states. Xue~\textit{et al.}~\cite{xue2012novel} describe a filtering approach for averaging measurements from multiple noisy gyroscopes. 

Eckenhoff~\textit{et al.}~\cite{eckenhoff2019sensor} describe a technique aimed at the fight against IMU sensor failure by employing switching between two IMU sensors. The technique is based on the multi-state constraint Kalman filter (MSCKF).
Our work is not directly solving the failure detection or fault-tolerance problem, although our method can automatically solve this issue due to the core idea of dynamically choosing the best sensors' axes, and there may be many resemblances on the final aim of our approach where we choose the best axes and discard the rest.

It can be declared that the pure probabilistic model of IMU measurements utilizes only a limited number from a wide range of parameters. 
In contrast to publications on the topic, including papers mentioned above that consider a purely probabilistic IMU model, Wang~\textit{et al.}~\cite{wang2019optimized} examine an extended single IMU non-linear calibration model to mimic real sensors better. 
For a reference of diverse IMU models and parameters, see \cite{titterton2004strapdown}.

One might need a high-quality IMU sensor (e.g., fiber optic gyroscope and high-grade MEMS accelerometer~\cite{deppe2017mems,narasimhan2015micromachined}) to achieve high accuracy of state estimation or could improve the quality of the IMU data by combining several cheap low-grade IMU sensors. 
While the majority of the above methods employ a high number of sensors (e.g., Skog~\textit{et al.}~\cite{skog2016inertial} use 32 cheap IMUs to achieve an accurate state estimation), we, in contrast, propose to use any number of IMUs and even with a low number (2-3) we achieve better accuracy of state estimation in comparison to averaging approaches. 
We explain this gain by direct handling of the systematic error while other methods use a large number of sensors relying on the purely probabilistic nature of the error.
We proceed the idea of our approach of BAC-gyroscope introduced in~\cite{faizullin2021best} and extend it to accelerometers to construct a complete BAC approach for accurate pose estimation.
\section{Background}\label{sec_background}
\subsection{State Variables and Frames of Reference}\label{sec_frames}

In this paper, we use the following frames of reference: the World frame denoted by $\{W\}$, the IMU frame $\{I\}$ or $\{I_i\}$ for each $i$-th IMU in the set of IMUs and the Master frame $\{M\}$ corresponding to an exteroceptive sensor (e.g., visual-camera, depth-camera, lidar, motion capture system markers, etc.), which is rigidly connected to the frame $\{I\}$. 

Rotation matrices $\lfR{\text{Target}}\rf{\text{Origin}} \in SO(3)$  represent the change of the reference frames: from the $\{\text{Origin}\}$ frame denoted as the right-hand subscript to the $\{\text{Target}\}$ frame denoted as the left-hand superscript.  In addition, by rotation matrices we express 3D orientations. 

Positions are expressed as 3D-vectors $\lf{\text{Frame}}\vect{p}~\in~\mathbb{R}^3$ in the $\{\text{Frame}\}$ frame, as well as  velocities and accelerations.
We also note vectors in the $\{\text{Target}\}$ reference frame pointing to the origin of coordinates of the $\{\text{Origin}\}$ reference frame by the following compact expression $\lf{\text{Target}}\vect{p}\rf{\text{Origin}}$.

\begin{figure}[t]
\includegraphics[width=\columnwidth]{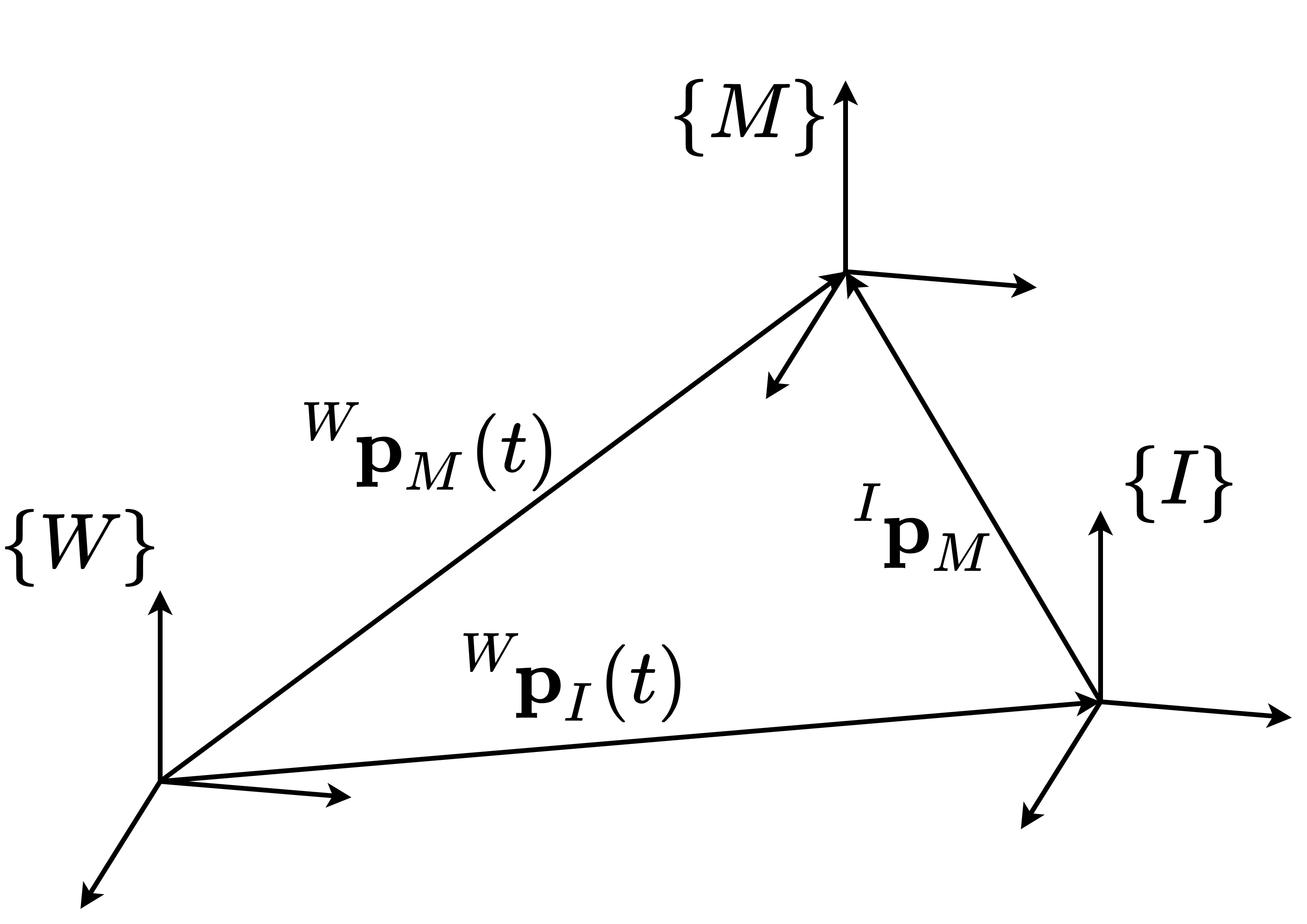}
\caption{Relation between the frames  $\{W\}$, $\{I\}$, $\{M\}$.
$\lf{W}\vect{p}\rf{I}(t)$~-~coordinates of $\{I\}$ in $\{W\}$.
$\lf{W}\vect{p}\rf{M}(t)$~-~coordinates of $\{M\}$ in $\{W\}$. 
$\lf{I}\vect{p}\rf{M}   $~-~coordinates of $\{M\}$ in $\{I\}$ that is time-invariant due to rigid connection between $\{I\}$ and $\{M\}$. Positions of the frames correspond to positions of their origins.}
\label{fig_frames}
\end{figure}

Fig.~\ref{fig_frames} depicts relation between the frames $\{W\}$, $\{I\}$, $\{M\}$ used in this work. 
Throughout the paper we use the following properties of rigid-body transformations~\cite{kevin2017modern}:
$$\lfR{\text{Target}}^{-1}\rf{\text{Origin}} = \lfR{\text{Target}}^\top\rf{\text{Origin}} = \lfR{\text{Origin}}\rf{\text{Target}},$$
$$\lf{\text{Origin}}\vect{p}\rf{\text{Target}} = - \lfR{\text{Origin}}\rf{\text{Target}} \lf{\text{Target}}\vect{p}\rf{\text{Origin}}.$$

Angular velocity $\lf{(\cdot)}\vect{\omega}~\in~\mathbb{R}^3$ of a reference frame and a linear acceleration $\lf{(\cdot)}\vect{a}~\in~\mathbb{R}^3$ of its origin are expressed by the coordinates of this frame and thus, for notation of the frame, we use a left-hand superscript: $\lf{I}\vect{\omega}$, $\lf{I}\vect{a}$, in this case, the IMU frame $\{I\}$. 

Orientations $R(t)$ and positions $\vect{p}(t)$ express continuous-time (c-t) quantities, while $R(t_k)$ and $\vect{p}(t_k)$ are discrete-time (d-t) sequences of orientations and positions for time $t_k$ and index $k$. The same notation is used for other variables. 

\subsection{IMU Model}\label{sec_imu}

Measurements of high-end factory-calibrated MEMS IMUs can be modeled as~in \cite{nikolic2016characterisation}:
\begin{equation}\label{eq_perf_gyro_cont}
    \begin{aligned}
        \lf{I}\tilde{\vect{\omega}}(t) &= \lf{I}\vect{\omega}(t) + \vect{b}_g(t) + \vect{\eta}_g(t),\\
         \lf{I}\tilde{\vect{a}}(t) &= \lf{I}\vect{a}(t) + \vect{b}_a(t) + \vect{\eta}_a(t) + \lfR{I}\rf{W}(t) \,\vect{g},
    \end{aligned}
\end{equation}
where $\lf{I}\tilde{\vect{\omega}}(t)$ and  $\lf{I}\tilde{\vect{a}}(t)$ are the measured values by the sensors,  $\lf{I}\vect{\omega}(t)$ and $\lf{I}\vect{a}(t)$ are the true angular velocity and linear acceleration, and $\vect{\eta}_g(t)$ and $\vect{\eta}_a(t)$ are Wiener processes. Vector $\vect{g}$ is the gravitational acceleration that is expressed by coordinates of $\{I\}$ by $\lfR{I}\rf{W}$.
The terms $\vect{b}_g(t)$ and $\vect{b}_a(t)$ are the slow varying continuous-time biases modeled as
\begin{equation}\label{eq_ct_bias}
    \begin{aligned}
        \dot{\vect{b}}_g(t) = - \frac{1}{\tau_{g}} \vect{b}_g(t) + \vect{\epsilon}_g(t),\\
         \dot{\vect{b}}_a(t) = - \frac{1}{\tau_{a}} \vect{b}_a(t) + \vect{\epsilon}_a(t),
    \end{aligned}
\end{equation}
where $\vect{\epsilon}_g(t)$ and $\vect{\epsilon}_a(t)$ are Wiener processes, and
$\tau_{b_g}$ and $\tau_{b_a}$ are correlation times of the biases~\cite{crassidis2006sigma}.

This model can be further improved by the introduction of correction lower triangular matrices $C_g$ and $C_a$ to take into account the non-unit scale of measurements and misalignment of the axes:
\begin{equation}\label{eq_cor_gyro_cont}
    \begin{aligned}
        \lf{I}\vect{\omega}(t) &= C_g \lf{I}\tilde{\vect{\omega}}(t) - \vect{b}_g(t) - \vect{\eta}_g(t),\\
        \lf{I}\vect{a}(t) &= 
        C_a \lf{I}\tilde{\vect{a}}(t) - \vect{b}_a(t) - \vect{\eta}_a(t) - \lfR{I}\rf{W}(t) \,\vect{g}.
    \end{aligned}
\end{equation}
In addition, our model employ non-unit relative orientation between accelerometer and gyroscope~\cite{rehder2016extending}; however, we do not include this parameter to keep notation clear for the reader. The acceleration influence on the gyroscope model is not included because of its insignificance. It is also assumed that IMU-sensor constitutes by itself a single-point sensor 
within one IMU sensor package. For additional discussion on this, please see \cite{skog2006calibration,rehder2016extending}.

The discrete-time model of (\ref{eq_cor_gyro_cont})  accordingly is
\begin{equation}\label{eq_cor_gyro_dis}
    \begin{aligned}
        \lf{I}\vect{\omega}(t_k) &= C_g \lf{I}\tilde{\vect{\omega}}(t_k) - \vect{b}_g(t_k) - \vect{n}_g(t_k),\\
        \lf{I}\vect{a}(t_k) &=  
        C_a \lf{I}\tilde{\vect{a}}(t_k) - \vect{b}_a(t_k) - \vect{n}_a(t_k) - \lfR{I}\rf{W}(t_k) \,\vect{g},
    \end{aligned}
\end{equation}
where the continuous-time biases are now  expressed by the discrete-time equations
\begin{equation}\label{eq_bias_dis}
    \begin{aligned}
        \Delta \vect{b}_g(t_k) = - \gamma_g \vect{b}_g(t_k) + \vect{\varepsilon}_{g}(t_k),\\
        \Delta \vect{b}_a(t_k) = - \gamma_a \vect{b}_a(t_k) + \vect{\varepsilon}_{a}(t_k),
    \end{aligned}
\end{equation}
$\Delta \vect{b}_g(t_k)$ and $\Delta \vect{b}_a(t_k)$ are the differences of biases:
\begin{equation}
    \begin{aligned}
        \Delta \vect{b}_g(t_k) &\triangleq \vect{b}_g(t_{k+1}) - \vect{b}_g(t_k),\\
        \Delta \vect{b}_a(t_k) &\triangleq \vect{b}_a(t_{k+1}) - \vect{b}_a(t_k).
    \end{aligned}
\end{equation}
The parameters $\gamma_g$ and $\gamma_a$ are dimensionless coefficients that take into account bias correlation time as described in the appendix of~\cite{crassidis2006sigma}. The random variables are
          $\vect{n}_g \sim \mathcal{N}(\vect{0}, \Sigma_{\vect{n          }_g})$, 
$\vect{\varepsilon}_g \sim \mathcal{N}(\vect{0}, \Sigma_{\vect{\varepsilon}_g})$, 
          $\vect{n}_a \sim \mathcal{N}(\vect{0}, \Sigma_{\vect{n          }_a})$, and 
$\vect{\varepsilon}_a \sim \mathcal{N}(\vect{0}, \Sigma_{\vect{\varepsilon}_a})$.

In this paper, we have chosen a fairly complete IMU model, although more complex alternatives are available in the literature \cite{titterton2004strapdown}.
The aim of this work, however, is to accept the limitations of the IMU model, where the systematic error is unavoidable, especially for mass-production MEMS IMUs, and to adapt our algorithm accordingly.
The nature of these errors is out of the scope of this work; however, they may be related to non-linearities, non-Gaussian distributions of noise, additional non-stationary parameters, fabrication tolerances, temperature, discretization errors, external forces, power supply conditions, etc.

\subsection{Kinematic Model}\label{sec_kinem}

The continuous-time kinematic model for obtaining orientation, velocity and position that is used in this work, as proposed in \cite{kevin2017modern}, is 
\begin{equation}\label{eq_corrected_imu}
    \begin{aligned}
        \lfdotR{W}\rf{I}(t) &= \lfR{W}\rf{I}(t) \cdot [\lf{I}\vect{\omega}]_\times(t),\\
        \lf{W}\dot{\vect{v}}\rf{I}(t) &=  \lfR{W}\rf{I}(t)\, \lf{I}\vect{a}(t),\\
        \lf{W}\dot{\vect{p}}\rf{I}(t) &=  \lf{W}\vect{v}\rf{I}(t),
    \end{aligned}
\end{equation}
where 
$[\vect{\omega}]_\times$ is a skew symmetric matrix:
\begin{align}
    [\vect{\omega}]_\times = \begin{bmatrix}  \omega^1\\ \omega^2\\ \omega^3 \end{bmatrix}_\times = \begin{bmatrix}  0 & -\omega^3 & \omega^2\\ \omega^3 & 0 & -\omega^1\\ -\omega^2 & \omega^1 & 0 \end{bmatrix}.
\end{align}

In order to obtain the discrete-time model, we make the following assumption.

{\bf Assumption 1}: {\em Angular velocities $\vect{\omega}$ and linear accelerations $\vect{a}$ are piece-wise constant functions over time and their values remain constant in any time interval $[t_k, t_{k+1}]$}.

{Assumption 1 is introduced intentionally, for fair comparison with AVE, in order to follow the same discrete-time integration method as in AVE competitor that in turn, is based on the method explained in~\cite{forster2016manifold}.
Alternative models consider continuous-time integration of interpolated measurements by high-order polynomials~\cite{furgale2012continuous}; however, these kinds of approaches are neither exempt from error.

Accordingly, the integration of the angular velocity $\lf{I}\vect{\omega}({t_k})$ and the double-integration of the linear acceleration $\lf{I}\vect{a}({t_k})$ are exactly calculated by
%
\begin{equation}\label{eq_irots_dis}
    \begin{aligned}
        \lfR{W}\rf{I}({t_{k+1}}) &= \lfR{W}\rf{I}(t_{k}) \cdot \Exp\left(\lf{I}\vect{\omega}({t_k})\cdot \Delta t_k\right),\\
        \lf{W}\vect{v}\rf{I}({t_{k+1}}) &= \lf{W}\vect{v}\rf{I}(t_{k}) + \lfR{W}\rf{I}(t_k)\, \lf{I}\vect{a}(t_k) \Delta t_k,\\
        \lf{W}\vect{p}\rf{I}({t_{k+1}}) &= \lf{W}\vect{p}\rf{I}(t_{k}) + \lf{W}\vect{v}\rf{I}(t_{k}) \Delta t_k + \frac{1}{2}\lfR{W}\rf{I}(t_k)\, \lf{I}\vect{a}(t_k) \Delta t_k^2,
    \end{aligned}
\end{equation}
where $\Delta t_k = t_{k+1}-t_k$. The function $\Exp(\vect{\theta})$ is a mapping from the tangent space of rotations $\mathbb{R}^3$ (spanning the Lie algebra $\mathfrak{so}(3)$) to the group of matrix rotations $SO(3)$. The inverse operation exists, the logarithm $\Ln:~SO(3)~\to~\mathbb{R}^3$.
The topic of Lie algebra is well documented, and it is a convenient way to express rotations or rigid body transformations \cite{kevin2017modern, barfoot2017}.

The pose of the Master frame $\{M\}$, given any of the $i$-th IMUs can be computed by similar equations:
\begin{equation}\label{eq_mrots_dis}
    \begin{aligned}
    \lfR{W}^i\rf{M}({t_{k+1}}) &= \lfR{W}^i\rf{M}(t_{k}) \cdot \Exp\left(\lf{M}\vect{\omega}({t_k})\cdot \Delta t_k\right),\\
     \lf{W}\vect{v}\rf{M}({t_{k+1}}) &= \lf{W}\vect{v}\rf{M}(t_{k}) + \lfR{W}\rf{M}(t_k) \lf{M}\vect{a}(t_k) \Delta t_k,\\
     \lf{W}\vect{p}\rf{M}({t_{k+1}}) &= \lf{W}\vect{p}\rf{M}(t_{k}) + \lf{W}\vect{v}\rf{M}(t_{k}) \Delta t_k\\ 
    &+ \frac{1}{2}\lfR{W}\rf{M}(t_k) \lf{M}\vect{a}(t_k) \Delta t_k^2,
    \end{aligned}
\end{equation}
where the angular velocity and linear acceleration are expressed via angular velocity and linear acceleration of some of IMUs by 
\begin{align}
    \lf{M}\vect{\omega} &= \lfR{M}\rf{I_i}\lf{I_i}\vect{\omega},\label{eq_omega_rot}\\
     \lf{M}\vect{a} &=   \lfR{M}\rf{I_i} \big( \lf{I_i}\vect{a} + \lf{I_i}\vect{a}_{inertial | M} \big).\label{eq_a_rot}
\end{align}

In turn, $\lf{I_i}\vect{a}_{inertial | M}$ is an inertial (also fictitious or pseudo) acceleration in the origin of $\{M\}$ additional to $\lf{I_i}\vect{a}$ that consists of the centrifugal and the Euler accelerations when rotating the rigidly attached frames of reference with $\lf{I_i}\vect{\omega}$~\cite{skog2016inertial}:
\begin{align}
    \lf{I_i}\vect{a}_{inertial | M} = \big( [\lf{I_i}\vect{\omega}]_\times^2 + [\lf{I_i}\vect{\dot \omega}]_\times \big) \lf{I_i}\vect{p}\rf{M},
\end{align}
where $\vect{\dot \omega}(t_k)$ we define as $\vect{\dot \omega}(t_k) = \frac{ \vect{\omega}(t_k) - \vect{\omega}(t_{k-1}) }{ \Delta t_{k-1} }$. Fig.~\ref{fig_a_inertial} depicts relation~(\ref{eq_a_rot}) for our configuration of sensors.

\begin{figure}[t]
\includegraphics[width=\columnwidth]{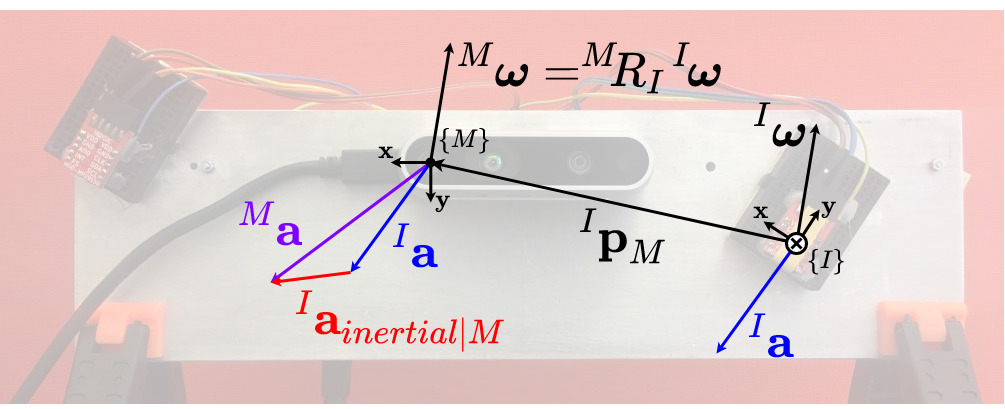}
\caption{Dependence between $\lf{I}\vect{a}$ and $\lf{M}\vect{a}$ through $\lf{I}\vect{a}_{inertial | M}$ for our configuration of sensors. 
$\{I\}$ is the frame connected to an IMU, $\{M\}$ is the Master frame, which coincides with a grayscale camera sensor frame. 
According to our notation, $\lf{I}\vect{\omega}$, $\lf{I}\vect{a}$ and $\lf{I}\vect{a}_{inertial | M}$ are expressed via coordinates in $\{I\}$; $\lf{M}\vect{\omega}$ and $\lf{M}\vect{a}$ has coordinates in $\{M\}$.
$\bullet$ and $\otimes$ indicate directions of $\vect{z}$-axis of every frame.}
\label{fig_a_inertial}
\end{figure}

\subsection{Averaged Virtual Estimator}\label{sub_baseline}

Given multiple gyroscope sensor readings, one can estimate the overall pose by fusing all available data into virtual measurements:
\begin{equation}\label{eq_baseline}
    \begin{aligned}
        \lf{V}\hat{\vect{\omega}} &= \frac{1}{N}\sum_{i=1}^N \lfR{V}\rf{I_i} \left(\lf{I_i}\tilde{\vect{\omega}} - \hat{\vect{b}}_{g_i}\right),\\
        \lf{V}\hat{\vect{a}} &=  \frac{1}{N}\sum_{i=1}^N \lfR{V}\rf{I_i} \left(\lf{I_i}\tilde{\vect{a}} - \hat{\vect{b}}_{a_i} +  \lf{I_i}\vect{a}_{inertial | V}\right),
    \end{aligned}
\end{equation}
under assumption of independent noise processes with the same parameters as in~(\ref{eq_perf_gyro_cont})~and~(\ref{eq_ct_bias}), matrices $C_g$, $C_a$ are not shown. $N$ is the number of IMUs, $\lfR{V}\rf{I_i}$ and $\lf{I_i}\vect{a}_{inertial | V}$ are the rotations and inertial accelerations to transform IMU measurements to the virtual frame $\{V\}$.

We have chosen to compare against one variant of the {\em virtual} IMU approach \cite{zhang2019lightweight}.
The authors utilize a probabilistic version of (\ref{eq_perf_gyro_cont}).
They report that virtual IMU measurements constructed by the weighted sum of measurements from multiple IMUs give better accuracy compared to single-IMU measurements. The authors also consider identical IMUs that simplify the measurement model to the expectation of measurements. We refer to the approach in (\ref{eq_baseline}) as the  Averaged Virtual Estimator (AVE) and will use this model as the \textit{baseline} for the comparison with our proposed model.

\section{Best Axes Composition for Multiple IMU Sensor Fusion in Open Loop}\label{sec_approach}

If the proposed IMU model (\ref{eq_cor_gyro_dis}) truthfully supports real data, that is, the statistical error has been accurately estimated, then the more data we use, the more accurate will be the estimator, ultimately leading to an averaged estimator as in (\ref{eq_baseline}) or similar approaches.

On the contrary, one can argue that the previous statement does not always hold, especially for mass-production IMUs, and we must make a different assumption.

{\bf Assumption 2: }{\em There exists a non-negligible source of the error, which can not be quantified in the current model (systematic uncertainty), which affects each axis of each IMU differently at each instant of time.}

One could calculate the statistics of this error given ground truth information. However, this will lead to more complex models, and still, they will not be exempt from systematic error.

We propose a different approach to fuse sensor data, which is based on the following Hypothesis. 

{\bf Hypothesis: }{\em The systematic error at each axis can not be predicted; however, it can be evaluated by the error in the estimated state. This error is non-stationary but does not change significantly for short time intervals.}

Based on this, we propose to choose dynamically those IMU axes, which have presented minimum error during evaluation in a method called {\em Best Axes Composition} (BAC).
In Sec.~\ref{sec_exp}, we show empirically the validity of the Hypothesis for the dataset configuration used.

We propose three stages or modes of working: 
(I) stationary and (II) time-dependent IMU parameters estimation when reference is observed (closed loop); (III) open-loop state estimation with no observations from reference. 

Stage I is aimed at intrinsic and extrinsic calibration of all IMUs. Stage II is the initial part of the main algorithm. It models the case when Master-based data are available and exploited as the visual data based estimator of the pose. In this stage the choice of the best axes is performed. Simultaneously, biases of all the IMUs are estimated. Stage III models the loss of Master-based pose estimation when the algorithm relies only on chosen pieces of inertial data for pose estimation.

\subsection{Stationary IMU Parameters Estimation}\label{sec_static_optim}

In order to evaluate all the available IMUs, we require an exteroceptive sensor (at the Master frame $\{M\}$) to provide ``ground truth'' measurements of absolute orientation compared with the estimated orientation from each IMU.

In the first preliminary stage we perform batch optimization for estimating the extrinsic and intrinsic parameters of IMUs, namely scale-misalignment matrices $C_{i_g}$ and $C_{i_a}$ introduced in~(\ref{eq_cor_gyro_cont}) and rotations of IMUs w.r.t.~Master $\lfR{M}\rf{I_i}$. Time-depended biases and poses are also estimated. This stage is additionally discussed in Sec.~\ref{sec_exp}.
States and parameters to be estimated are
\begin{equation}
    \begin{aligned}
        \Theta_i = \big\{
            &\Ln(\lfR{W}^{i}\rf{M}(t_k)),\: 
            \lf{W}\vect{v}^{i}\rf{M}(t_k),\: 
            \lf{W}\vect{p}^{i}\rf{M}(t_k),\\ 
            &\vect{b}_g^i(t_k),\: 
            \vect{b}_a^i(t_k),\: 
            \Ln(\lfR{M}\rf{I_i}),\:
            C_{i_g},\: 
            C_{i_a} 
        \big\}.
    \end{aligned}
\end{equation}
The number of the variables is $(15K + 15)N$ for $N$ IMU sensors and $K$ timestamps.

Assuming eqs.~(\ref{eq_cor_gyro_dis}),~(\ref{eq_bias_dis}),~(\ref{eq_mrots_dis}),~(\ref{eq_omega_rot}), (\ref{eq_a_rot}), the cost function given the values is the following for each IMU: 
\begin{equation}\label{eq_loss}
    \begin{aligned}
        \mathcal{L}_i(\Theta_i) =& \sum_{k=1}^K \left\| \Ln (\lfR{M}^{\scriptscriptstyle GT}\rf{W}(t_k) \lfR{W}^i\rf{M}(t_k)) \right\|_{\Sigma_{\theta}}^2\\
        +& \sum_{k=1}^K \left\| \lf{W}\vect{v}^{\scriptscriptstyle GT}\rf{M}(t_k) - \lf{W}\vect{v}^{i}\rf{M}(t_k) \right\|_{\Sigma_{v}}^2\\
        +& \sum_{k=1}^K \left\| \lf{W}\vect{p}^{\scriptscriptstyle GT}\rf{M}(t_k) - \lf{W}\vect{p}^{i}\rf{M}(t_k) \right\|_{\Sigma_{p}}^2\\
        +& \sum_{k=1}^{K-1} \left\| \Delta \vect{b}_g^i(t_k) + \gamma_g \vect{b}_g^i(t_k)\right\|_{\Sigma_{\vect{n}_{b_g}}}^2\\
        +&  \sum_{k=1}^{K-1} \left\| \Delta \vect{b}_a^i(t_k) + \gamma_a \vect{b}_a^i(t_k)\right\|_{\Sigma_{\vect{n}_{b_a}}}^2, 
    \end{aligned}
\end{equation}
where the first three terms are the errors in orientation, velocity, and position, and the last two terms are conditions to biases. Further, $\lfR{M}^{\scriptscriptstyle GT}\rf{W}(t_k)$, $\lf{W}\vect{v}^{\scriptscriptstyle GT}\rf{M}(t_k)$, and $\lf{W}\vect{p}^{\scriptscriptstyle GT}\rf{M}(t_k)$ are ground truth data provided by Master sensor. $\lfR{W}^i\rf{M}(t_k)$ is orientation estimate, $\lf{W}\vect{v}^{i}\rf{M}(t_k)$ and $\lf{W}\vect{p}^{i}\rf{M}(t_k)$ are velocity and position estimates.  
$\Sigma_\theta$, $\Sigma_v$ and $\Sigma_p$ are covariance matrices described in~\cite{forster2015supplementary}~and~\cite{sola2018micro}.

The final cost of all IMU measurements to be minimized is
\begin{align}
    \mathcal{L}(\Theta) =& \sum_{i=0}^N \mathcal{L}_i(\Theta_i) \to \min_{\Theta}.
    \label{eq_final_loss}
\end{align}

\subsection{Time-dependent IMU Parameters Estimation}

The second stage is the initial part of the main algorithm. Again, we treat observations from the Master sensor as ground truth poses. It includes further continuous optimization of the cost in (\ref{eq_final_loss}) but with 
only states to be estimated: 
$\Theta_i~=~ \{
    \Ln(\lfR{W}^{i}\rf{M}(t_k)),  
    \lf{W}\vect{v}^{i}\rf{M}(t_k), 
    \lf{W}\vect{p}^{i}\rf{M}(t_k), 
    \vect{b}_g^i(t_k),  
    \vect{b}_a^i(t_k) 
\}$, $15KN$ values in total. Here we use formerly estimated scale-misalignment and rotation parameters for every IMU. 

Estimation of biases in this stage is followed by the best axes choice for the third stage of the algorithm, where pose estimation is performed in open loop without the Master's aid.

\subsection{Best Axes Composition}\label{sec_bac}

Our approach, the Best Axes Composition (BAC), performs choice of three non-coplanar gyroscope axes and three non-coplanar accelerometer axes to be utilized from the whole set of $3N+3N$ gyroscope plus accelerometer axes for open-loop pose estimation instead of averaging, as described in~\ref{sub_baseline}. Other axes are ignored for data fusion since they are worse fitted, and thus, the systematic error related to these axes is bigger. The criterion for the BAC to choose a specific axis among the set of the homonymous (but not necessarily collinear) axes of different IMUs is \textit{the lowest estimated mean squared error within the last $p$ Master aided measurements}. 

The BAC consists of two phases: BAC-gyroscope and BAC-accelerometer, since the second phase utilizes the results of the first one.
Below is the description of the phases.

\subsubsection{BAC-gyroscope}

There are $p$ last reference $\lfR{W}^{\scriptscriptstyle GT}\rf{M}(t_k)$ and estimated $\lfR{W}^{\scriptscriptstyle M}\rf{M}(t_k)$ orientations of the Master of the second stage. 
The orientation error for IMU $i$ and timestamp $k$ expressed in the respective IMU frame~$i$ is 
\begin{equation}
    \begin{aligned}
        \vect{e}_{i}(t_k) =& \Ln\left(\left( \lfR{W}^{\scriptscriptstyle GT}\rf{M}(t_k) \lfR{M}\rf{I_i}\right)^\top \lfR{W}^i\rf{M}(t_k) \lfR{M}\rf{I_i}\right)\\
        =& \Ln\left( \lfR{I_i}\rf{M} \lfR{M}^{\scriptscriptstyle GT}\rf{W}(t_k) \lfR{W}^i\rf{M}(t_k) \lfR{M}\rf{I_i} \right). 
    \end{aligned}
\end{equation}
Here, the mapping of the error in the Master to the IMU frames has been performed.
The mapping allows the algorithm to capture the contribution of measurements of every single axis of a gyroscope to a three-dimensional error vector 
\begin{align}\label{eq_error}
    \vect{e}_{i} = [e^x_i, e^y_i, e^z_i]^\top.
\end{align}

Then we choose the best axis from the homonymous axes of the IMUs. This is done by MSE criterion of the sets of errors $e^\alpha_i$ in time instances $t_{K-p}$,~$t_{K-p+1}$,~\dots,~$t_{K}$ for every axis $\alpha~\in \{x,y,z\}$ 
with the assumption that chosen axes are not coplanar:
\begin{align}
    i_{best}^\alpha = \arg \min_i \sum_{k=K-p}^K | e^\alpha_i (t_k)|^2.
    \label{eq_vot_tule}
\end{align}

We do it three times for every axis to compose axes of the virtual gyroscope. Thus, on this step the set of three non-coplanar axes 
\begin{align}\label{eq_best_axes}
    \{i_{best\:gyro}^x, i_{best\:gyro}^y, i_{best\:gyro}^z\}
\end{align}
from the set of IMU sensors for open-loop estimation have been chosen by criterion (\ref{eq_vot_tule}). 

The next step is the final composition of the axes and their measurements to assemble accurate distributed single virtual IMU measurements. We match axes of the virtual IMU with the Master frame for simplicity, but there are no limitations on the position and orientation of the virtual IMU. 
For this, consider now a problem of expression of angular velocity in the Master frame via non-coplanar elements of angular velocities 
$\lf{I_i}\omega^x,~\lf{I_j}\omega^y$ and $\lf{I_k}\omega^z$ 
of arbitrary chosen IMUs~$i$,~$j$~and~$k$:
\begin{align}\label{eq_map}
    \lf{M}\vect{\omega} = A \cdot \begin{bmatrix} \lf{I_i}\omega^x\\ \lf{I_j}\omega^y\\ \lf{I_k}\omega^z \end{bmatrix},
\end{align}
where $A \in \mathbb{R}^{3\times3}$ is a linear mapping that is needed to be found. The matrix $A$ is non-orthonormal and does not belong to $SO(3)$ because the homonymous axes of different IMUs do not have to be coplanar in general. The matrix $A$ can be obtained by exploring element by element of the  vector
\begin{align}
    \lf{I_i}\vect{\omega} = \lfR{I_i}\rf{M}\lf{M}\vect{\omega},
\end{align}
which is the inverse relation in (\ref{eq_omega_rot}). This gives each component of the angular velocity as a dot product
\begin{align}\label{eq_comp}
    \lf{I_i}\omega^\alpha = \lf{I_i} \vect{r}^\alpha \rf{M} \lf{M}\vect{\omega},
\end{align}
for every axis $\alpha$ of $\lf{I_i}\vect{\omega}$, where the row vector $\vect{r}^\alpha \rf{M}$ is the $\alpha$-th row of $\lfR{I_i}\rf{M}$. 

By stacking all the velocity components we obtain
\begin{align}
    \begin{bmatrix} \lf{I_i}\omega^x\\ \lf{I_j}\omega^y\\ \lf{I_k}\omega^z \end{bmatrix} = \begin{bmatrix} \lf{I_i} \vect{r}^x \rf{M}\\ \lf{I_j} \vect{r}^y \rf{M}\\ \lf{I_k} \vect{r}^z \rf{M} \end{bmatrix} \lf{M}\vect{\omega},
\end{align}
and finally,
\begin{align}\label{eq_distr}
    \lf{M}\vect{\omega} = \begin{bmatrix} 
        \lf{I_i} \vect{r}^x \rf{M}\\ 
        \lf{I_j} \vect{r}^y \rf{M}\\ 
        \lf{I_k} \vect{r}^z \rf{M}
    \end{bmatrix}^{-1} \begin{bmatrix} \lf{I_i}\omega^x\\ \lf{I_j}\omega^y\\ \lf{I_k}\omega^z \end{bmatrix}
\end{align}
is the solution of~(\ref{eq_map}).

Changing superscripts $I_i$, $I_j$, $I_k$ in~(\ref{eq_distr}) on $I_{i_{best\:gyro}^x}$, $I_{i_{best\:gyro}^y}$, $I_{i_{best\:gyro}^z}$
accordingly we get BAC-gyroscope. 

\subsubsection{BAC-accelerometer}\label{sec_bac_acc}

BAC-accelerometer approach follows the same way. For $p$ last reference and estimated positions of the second stage, the position error for IMU $i$ and timestamp $k$ expressed in the respective IMU frame $i$ is
\begin{equation}\label{eq_errror_acc}
    \begin{aligned}
        \vect{e}_{i}(t_k) &= \lf{I_i}\vect{p}\rf{W}(t_k) - \lf{I_i}\vect{p}^{\scriptscriptstyle GT}\rf{W}(t_k)\\
        &= \lfR{I_i}\rf{M} \big( \lfR{M}^{\scriptscriptstyle BAC}\rf{W}(t_k) \lf{W}\vect{p}\rf{M}(t_k) - \lfR{M}^{\scriptscriptstyle GT}\rf{W}(t_k) \lf{W}\vect{p}^{\scriptscriptstyle GT}\rf{M}(t_k) \big),
    \end{aligned}
\end{equation}
where $\lfR{M}^{\scriptscriptstyle BAC}\rf{W}(t_k)$ is the orientation estimated employing the measurements of distributed gyroscope (\ref{eq_distr}). 

Then, by MSE criterion (\ref{eq_vot_tule}) applied to (\ref{eq_errror_acc}) we obtain the best axes but for distributed accelerometer now:
\begin{align}\label{eq_best_axes_acc}
    \{i_{best\:acc}^x, i_{best\:acc}^y, i_{best\:acc}^z\}.
\end{align}

The problem of expression of linear acceleration in the Master frame via non-coplanar elements of linear accelerations
$\lf{I_i}a^x,~\lf{I_j}a^y$ and $\lf{I_k}a^z$ 
of arbitrary chosen IMUs~$i$,~$j$~and~$k$ is expressed by:
\begin{align}\label{eq_map_acc}
    \lf{M}\vect{a} = B \cdot \begin{bmatrix} \lf{I_i}a^x + \lf{I_i} a_{inertial}^x\\ \lf{I_j}a^y + \lf{I_j} a_{inertial}^y\\ \lf{I_k}a^z + \lf{I_k} a_{inertial}^z \end{bmatrix},
\end{align}
where $B \in \mathbb{R}^{3\times3}$ to be found that has the same properties as $A$ in (\ref{eq_map}). The matrix $B$ can be obtained by exploring element by element of the vector
\begin{align}\label{eq_a_rot_inv}
    \lf{I_i}\vect{a} &= \lfR{I_i}\rf{M}\lf{M}\vect{a} - \lf{I_i}\vect{a}_{inertial},
\end{align}
which is the inverse relation in (\ref{eq_a_rot}). 
Here, the inertial acceleration is expressed now through the BAC-gyroscope angular velocity (\ref{eq_distr}) with a proper transformation to $\{I\}$:
\begin{align}
    \lf{I_i}\vect{a}_{inertial} 
    = \lfR{I_i}\rf{M} \big( [\lf{M}\vect{\omega}]_\times^2 + [\lf{M}\vect{\dot \omega}]_\times \big) \lfR{M}\rf{I_i} \lf{I_i}\vect{p}\rf{M},
\end{align}
where we employ $[ R \vect{\omega}]_\times = R [\vect{\omega}]_\times R^\top$~\cite{sola2018micro}.
Each component of the acceleration (\ref{eq_a_rot_inv}) is
\begin{align}\label{eq_comp_acc}
    \lf{I_i} a^\alpha = \lf{I_i} \vect{r}^\alpha \rf{M} \lf{M}\vect{a} - \lf{I_i} a_{inertial}^\alpha,
\end{align}
for every axis $\alpha$ of $\lf{I_i}\vect{a}$.

By stacking all the acceleration components we obtain
\begin{equation}
    \begin{aligned}
        \begin{bmatrix} \lf{I_i}a^x\\ \lf{I_j}a^y\\ \lf{I_k}a^z \end{bmatrix} = 
        \begin{bmatrix}
            \lf{I_i} \vect{r}^x \rf{M}\\
            \lf{I_j} \vect{r}^y \rf{M}\\
            \lf{I_k} \vect{r}^z \rf{M}
        \end{bmatrix} \lf{M}\vect{a} - 
        \begin{bmatrix} 
            \lf{I_i} a_{inertial}^x\\
            \lf{I_j} a_{inertial}^y\\
            \lf{I_k} a_{inertial}^z
        \end{bmatrix},
    \end{aligned}
\end{equation}

and finally,
\begin{equation}\label{eq_distr_acc}
    \begin{aligned}
        \lf{M}\vect{a} = 
        \begin{bmatrix}
            \lf{I_i} \vect{r}^x \rf{M}\\
            \lf{I_j} \vect{r}^y \rf{M}\\
            \lf{I_k} \vect{r}^z \rf{M}
        \end{bmatrix}^{-1}
        \begin{bmatrix} 
            \lf{I_i}a^x + \lf{I_i} a_{inertial}^x\\
            \lf{I_j}a^y + \lf{I_j} a_{inertial}^y\\
            \lf{I_k}a^z  +\lf{I_k} a_{inertial}^z
        \end{bmatrix}
    \end{aligned}
\end{equation}
is the solution of~(\ref{eq_map_acc}).

Changing superscripts $I_i$, $I_j$, $I_k$ in~(\ref{eq_distr_acc}) on $I_{i_{best\:acc}^x}$, $I_{i_{best\:acc}^y}$, $I_{i_{best\:acc}^z}$
accordingly we get BAC-accelerometer. 

As a result for the third stage, we have prepared accurate distributed IMU (\ref{eq_distr}), (\ref{eq_distr_acc}) for open-loop orientation and position estimation. 

\subsection{Open-loop}
The third stage of the approach is open-loop pose estimation by only IMUs measurements without Master aid. This is an integration of distributed IMU measurements (\ref{eq_distr}), (\ref{eq_distr_acc}) with constant biases equal to the last estimated value in the second stage of the approach: $\vect{b}_g = \vect{b}_g(t_K)$, $\vect{b}_a = \vect{b}_a(t_K)$. 

As mentioned in Sec.~\ref{sec_intro}, a highly dynamic motion of aerial robots or augmented and virtual reality systems require as low as possible processing time of data along with frequent updates of a current state (high data rate). Using only visual data requires several tens or hundreds of milliseconds delay between events and their estimation. In these circumstances, IMU sensors provide a much faster alternative, and BAC outperforms state-of-the-art IMU solutions in accuracy over short time horizons. 

\section{Implementation}\label{sec_impl}

\begin{figure}[t]
\frame{\includegraphics[width=\columnwidth]{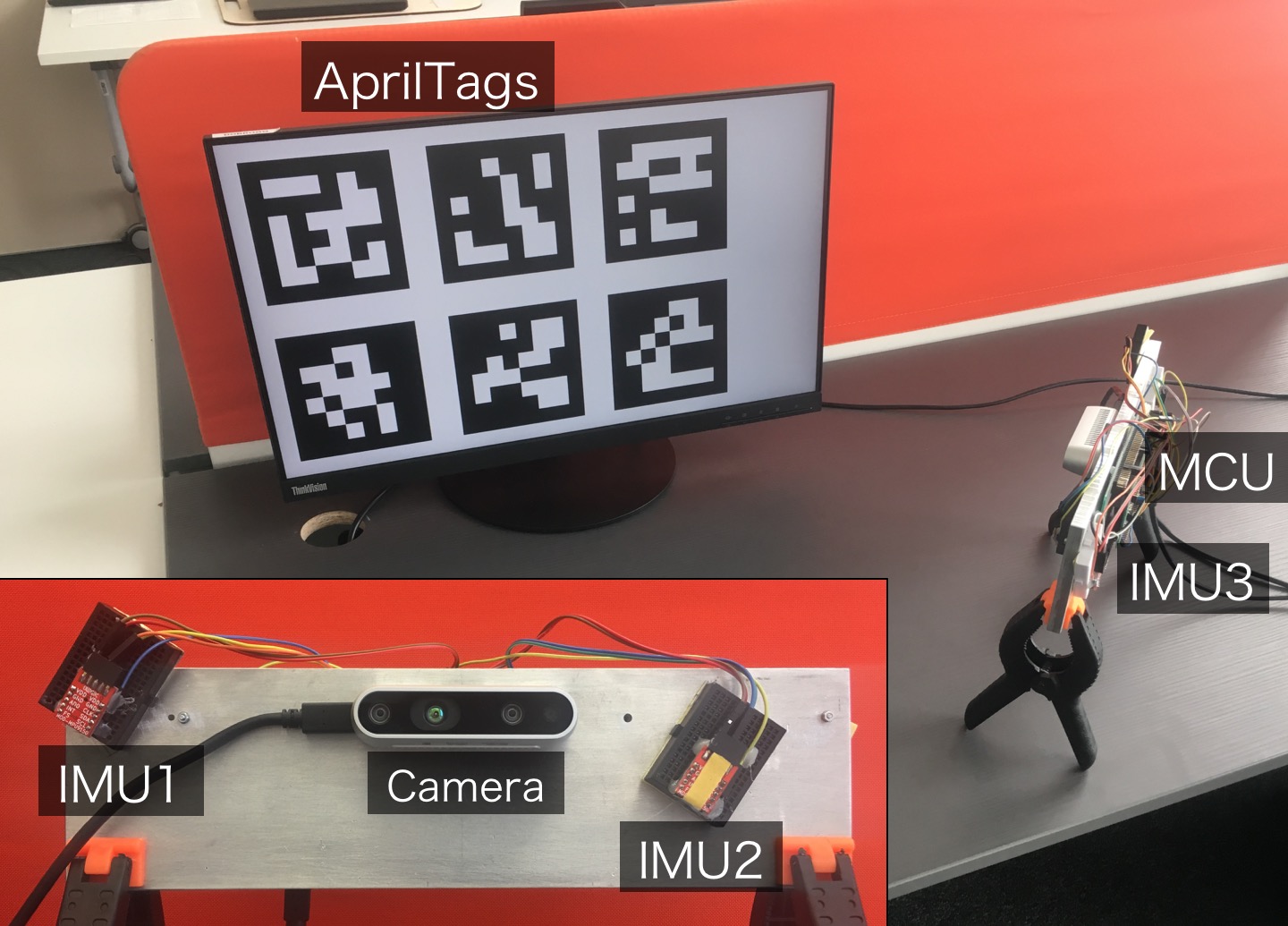}}
\caption{Setup for data collection. Camera and IMUs rigidly connected on aluminium platform. Camera provides ground truth orientation and position data by detecting pose of motionless AprilTags array displayed on the In-Plane Switching monitor. MCU prepares and transmits IMU data to laptop. AprilTags mean pose coincides with $\{W\}$, IMUi frame is $\{I_i\}$, Camera frame is $\{M\}$.}
\label{fig_sensor}
\end{figure}

In order to carry out experiments and evaluate our approach, we have developed a handheld visual-inertial sensor consisting of three IMUs MPU9150, Intel RealSense D435 depth camera (only one gray-scale global-shutter visual camera was used) calibrated beforehand, and a developers board based on the MCU STM32F051.
All the sensors are rigidly attached to a metal plate, as shown in Fig.~\ref{fig_sensor}.

Ground truth data are provided by tracking a custom-designed array of six AprilTags~\cite{olson2011apriltag}. This array has benefits over a single tag such that we are able to check for the mutual orientation and position error between the poses of every detected tag for robustness on the ground truth data generated. 
In addition, we use a screen to show the tags instead of a printed paper sheet to ensure that tags present no distortions, and they are still valid fiducial markers since dimensions are known by the pixel size of the monitor. The World frame $\{W\}$ coincides with the frame of this array of markers. The averaging to obtain orientation $\lfR{W}\rf{M}$ is performed by averaging in $SO(3)$~\cite{moakher2002means}.

The MCU performs data collection of IMU data. All the IMU sensors are synchronized on the hardware level and clocked by a shareable, highly accurate quartz clock generator. The sync of the sensors, in addition, is verified with sub-millisecond accuracy and precision by Python implementation\footnote{\url{https://pypi.org/project/twistnsync/}} of~\cite{faizullin2021twist}. 
The data rate of synchronized IMUs is 342 Hz. The frame rate of the camera is 30 fps with constant exposure time. Time synchronization Camera-IMUs is performed by correlation of absolute value of angular velocities, and has accuracy better than 3 ms.

IMUs are manually placed on the board, roughly maintaining the same orientations ($\pm 5^{\circ}$ for each axis at most). It is enough to avoid any singularities in the computation of matrices $A$ and $B$ explained in Sec.~\ref{sec_bac} since heterogeneous axes of any pair of IMUs are not coplanar.

Batch optimization is performed in PyTorch Python package by optimizing on the manifold of rotation matrices according to the technique described in~\cite{forster2016manifold}. The chosen optimizer is Adam with a variable learning rate 0.001-0.1 and 1400 epochs. 

Ground truth orientation and position were interpolated by a cubic spline to coincide with every IMU measurement. We have used the same prior covariance for each gyroscope axis, the same holds for accelerometer axes; although in the literature, some authors make a distinction regarding the z-axis \cite{nikolic2016characterisation}. For our IMUs, the correlation times of the biases $\tau_{g}$ and $\tau_{a}$ are of hundreds of seconds, this is in accordance with~\cite{nikolic2015maximum}. Taking these high values into account, we ignore terms with coefficients $\gamma_g$ and $\gamma_a$ in~(\ref{eq_bias_dis}).

\section{Experiments}\label{sec_exp}
We evaluate our approach on 44 non-overlapping tracks where each lasts 15 seconds. During the first 10 s of every track, pose estimation with Master aid is performed (Stage II), whereas, during the rest 5 s, the pose is estimated using only inertial data in open loop (Stage III).

Calibration, which is the first stage of the experiments, was carry out separately with a specific trajectory to gather as much diversity of poses as possible. It is important because common tracks have poor diversity to get accurate estimations for constant parameters of the sensor. 
This stage can also be done by other calibration tools, for instance, by~\cite{rehder2016extending}.

We consider a comparison with the competitors in short horizons of time that last 5 s for orientation estimation and 1.5 s for position estimation to highlight (i) the performance of BAC, (ii) its dynamic nature, and (ii) the need for proper treatment of this method. We state that the method greatly works right after the loss of complement sensor (camera, lidar, etc.) data and degrades in a higher horizon of time to MIMU or single IMU case and needs repetitive updates by aiding sensor data to outperform the competitors continuously.



We compare the averaged ratio (in percent) between the chosen baseline AVE and other estimation methods: single IMU and our proposed method in Fig.~\ref{fig_impr}. 
The higher the curve, the better accuracy of a method compared with the AVE, which is a flat constant line. It can be seen that for orientation estimation (Fig.~\ref{fig_impr_gyro}), our proposed BAC method outperforms the baseline significantly up to 2.5 sec in open loop and shows more than 10\% better performance within one second time horizon. For position estimation (Fig.~\ref{fig_impr_acc}), BAC outperforms the baseline up to 20\% and up to 0.4 sec in open loop.
\begin{figure*}[t]
     \centering
     \begin{subfigure}[b]{0.49\textwidth}
         \centering
         \includegraphics[width=\textwidth]{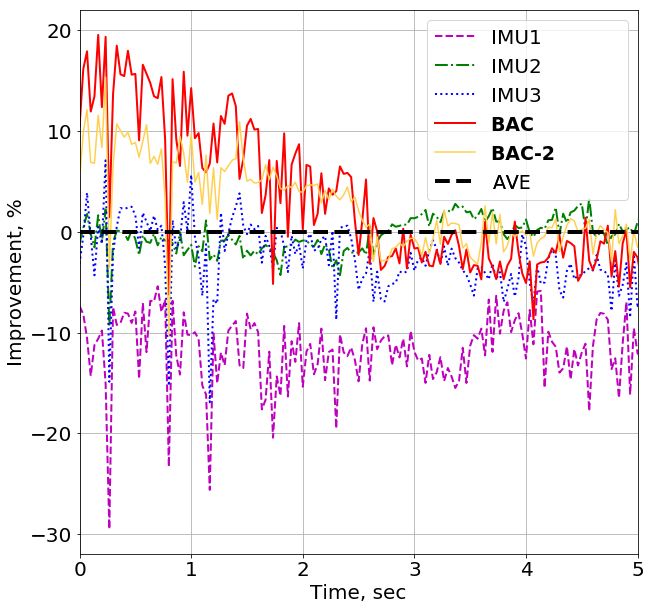}
         \caption{Orientation}
         \label{fig_impr_gyro}
     \end{subfigure}
     \hfill
     \begin{subfigure}[b]{0.49\textwidth}
         \centering
         \includegraphics[width=\textwidth]{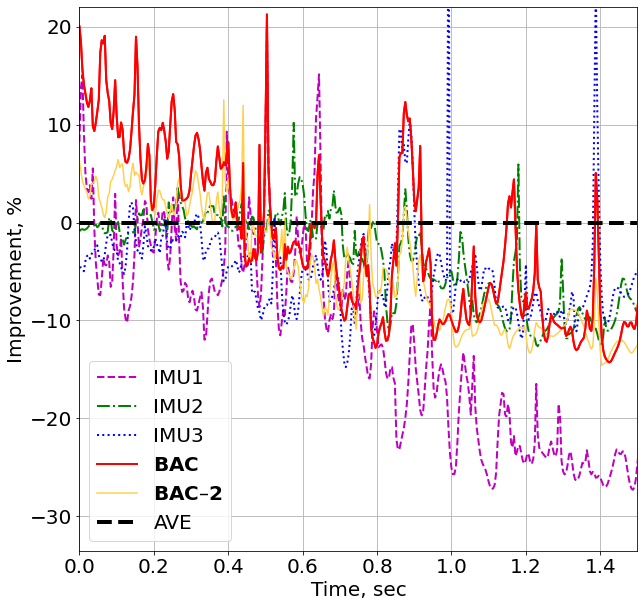}
         \caption{Position}
         \label{fig_impr_acc}
     \end{subfigure}
        \caption{Accuracy comparison for the stage III. The higher curve the better. Orientation and position accuracy improvement by our proposed method wrt. AVE averaging method in percent for 44 tracks averaged value of errors. The higher curve, the better accuracy. Single IMUs estimation accuracy also depicted.}
        \label{fig_impr}
\end{figure*}

The results also show the same performance of IMU2 and IMU3 of the sensor while poor performance of IMU1 on average in both orientation and position estimation. The poor performance of the position estimation by IMU1 is a consequence of non-accurate orientation results since the orientation estimate directly affects the velocity and position estimates in~(\ref{eq_mrots_dis}). In open-loop Stage III, the orientation errors cannot be resolved due to the absence of correcting information from the Master sensor.

We also apply our approach for two IMU sensors (BAC-2), namely IMU2 and IMU3. While accuracy for this setup is less than BAC using three IMUs, it is clear that it is also better than three IMU AVE within the same horizons above (for orientation estimation, it is more observable). This suggests the idea that even a less accurate sensor (IMU1 in our case) contributes well to overall improvement. 

Fig.~\ref{fig_freq} shows that the proposed method dynamically utilizes every axis of every gyroscope and every accelerometer during the experiments, confirming empirically that the nature of the error is non-stationary and different for each IMU.
\begin{figure}[t]
\includegraphics[width=\columnwidth]{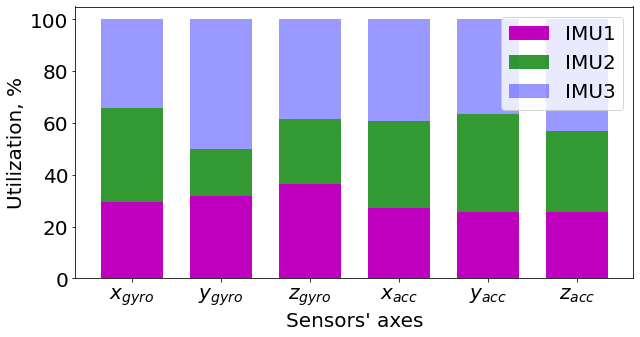}
\caption{Frequency of utilization of every axis of every gyroscope and accelerometer. This figure shows that the selection of the best axes is time-varying. 
}
\label{fig_freq}
\end{figure}

Fig.~\ref{fig_2-3} expresses the typical orientation error value of the second and third stages for different gyroscopes and methods. On the left top corner of the picture, error values for every axis of gyroscopes are depicted. There, dotted curves are chosen by our BAC criteria as more accurate in this case. We can see that these chosen axes for estimation of orientation in the third stage give better accuracy than every single IMU or AVE data fusion method, and the accuracy stays lower than 0.004 rad in 5 seconds time horizon. 
\begin{figure}[t]
\includegraphics[width=\columnwidth]{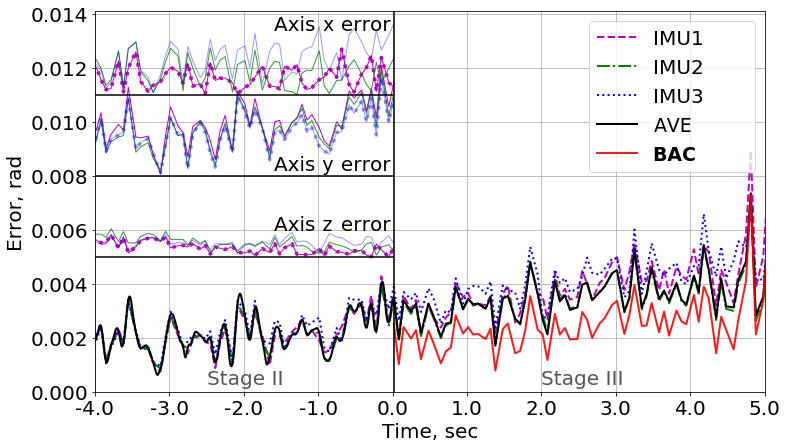}
\caption{Typical orientation estimation error with Master aid (Stage II) and in open-loop (Stage III). The top left part is the error in orientation measured for every axis of every IMU sensor in Stage II, dotted curves are chosen by the proposed algorithm as more accurate in this case for Stage III.}
\label{fig_2-3}
\end{figure}

Fig.~\ref{fig_2-3_acc} shows typical position error growth in the third, open-loop stage. The BAC outperforms AVE up to 0.4 seconds then becomes less accurate than AVE. We connect BAC-accelerometer degradation before AVE with a stronger influence of stochastic errors onto accelerometer data in contrast to the gyroscope case, where systematic errors contribute more. Another reason of that is the higher dynamic of systematic error fluctuations within a short time window. After 1 second of open loop, the errors for both AVE and BAC grow rapidly due to the double integration of noisy accelerometer measurements and growing error in orientation estimation ((\ref{eq_irots_dis}) or (\ref{eq_mrots_dis})). 

These results lead us to make the following conclusion. The lower bound of an exteroceptive Master sensor frame rate to gather data about the environment should be at 1-2 fps to allow an inertial sensor to keep pose estimation accuracy on the highest level. Lower frame rates are not recommended due to the fast error growth of position estimation by the solely inertial sensor. 
However, in case of only orientation estimation by an inertial sensor, discarding accelerometer measurements and consequently position estimation, orientation error growth dependence on time is close to linear relationship, and the frame rate lower bound may be decreased more. 
\begin{figure}[t]
\centering
\includegraphics[width=0.6\columnwidth]{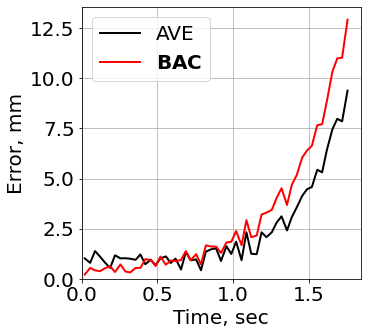}
\caption{Typical position error growth for open loop (Stage III). The BAC outperforms AVE up to 0.4 seconds then become less accurate than AVE. After 1 s in open loop, errors grow rapidly for both methods.}
\label{fig_2-3_acc}
\end{figure}

\section{Conclusions}\label{sec_concl}

We have proposed the Best Axes Composition (BAC) method of IMU sensors data fusion that takes into account systematic error by choosing three non-coplanar axes from gyroscopes and three non-coplanar axes from accelerometers of different IMU sensors. To do that, we have proposed and verified the Hypothesis (Sec.~\ref{sec_approach}), and as a result, our approach outperforms the Averaged Virtual Estimator (AVE) method up to several seconds for orientation and up to several hundreds of milliseconds for position estimation in open loop. 
The faster degradation of the position estimation improvement we have connected with the following reasons: (i) stronger influence of the stochastic errors on the accelerometers compared to the gyroscopes and (ii) systematic errors fluctuations with higher dynamic during lower time horizons. 

We have shown that our method requires only two IMUs to significantly improve orientation estimation, while for position estimation, the improvement is also considerable.

Taking into account the BAC accuracy degradation over time, we have provided the frame rate lower bounds of an exteroceptive sensor data for both orientation and full pose estimation to keep the accuracy of the method and any algorithm based on it on the highest level.












\bibliographystyle{bib/cas-model2-names}

\balance

\bibliography{bib/imu}

\bio{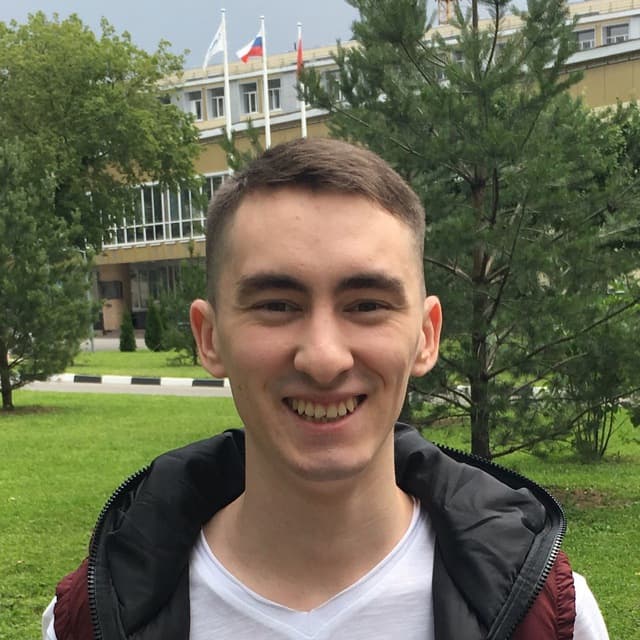}
\textbf{Marsel Faizullin} is finishing his Ph.D. study at Mobile Robotics Laboratory, Skolkovo Institute of Science and Technology, Russia. He received his B.S and M.S. from Moscow Institute of Science and Technology, Russia. His current research interests include robot perception, IMU data fusion, sensor synchronization and sensor networks. He is now working on industrial project on hardware-software data acquisition system.
\endbio

\bio{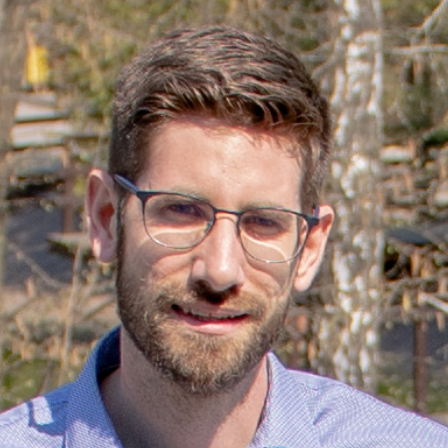}
\textbf{Gonzalo Ferrer} obtained  his Ph.D. in Robotics from the {\em Universitat Polit\`ecnica de Catalunya} (UPC), Barcelona, Spain in 2015 and worked during two years as a Research Fellow (postdoc) at the APRIL lab. in the department of Computer Science and Engineering at the University of Michigan. In 2018, Gonzalo  joined the Skolkovo Institute of Science and Technology as an Assistant Professor. He is heading the Mobile Robotics lab., focusing his research on planning, perception and how to combine both into new solutions in robotics.
\endbio

\end{document}